\documentclass[runningheads]{llncs}

 \usepackage{iciap}

\usepackage{iciapabbrv}

\usepackage{graphicx}
\usepackage{booktabs}
\usepackage{multicol}
\usepackage{multirow}
\usepackage{caption}
\usepackage{comment}
\usepackage[inkscape]{svg}
\usepackage{enumitem}
\usepackage[accsupp]{axessibility}  

\usepackage{hyperref}

\usepackage{orcidlink}

\usepackage[dvipsnames]{xcolor}

\begin{document}

\newcommand{\methname}{RoadFusion\xspace}
\title{RoadFusion: Latent Diffusion Model for Pavement Defect Detection}

\titlerunning{}

\author{Muhammad Aqeel\inst{}\orcidlink{0009-0000-5095-605X} \and 
Kidus Dagnaw Bellete \and
Francesco Setti\inst{}\orcidlink{0000-0002-0015-5534}} 

\authorrunning{M. Aqeel et al.}

\institute{Dept. of Engineering for Innovation Medicine, University of Verona\\ 
Strada le Grazie 15, Verona, Italy \\
Contact author: \email{muhammad.aqeel@univr.it}
}

\maketitle

\begin{abstract}


Pavement defect detection faces critical challenges including limited annotated data, domain shift between training and deployment environments, and high variability in defect appearances across different road conditions. We propose \methname, a framework that addresses these limitations through synthetic anomaly generation with dual-path feature adaptation. A latent diffusion model synthesizes diverse, realistic defects using text prompts and spatial masks, enabling effective training under data scarcity. Two separate feature adaptors specialize representations for normal and anomalous inputs, improving robustness to domain shift and defect variability. A lightweight discriminator learns to distinguish fine-grained defect patterns at the patch level. Evaluated on six benchmark datasets, \methname achieves consistently strong performance across both classification and localization tasks, setting new state-of-the-art in multiple metrics relevant to real-world road inspection.
\end{abstract}

\section{Introduction}
\label{sec:intro}

Road infrastructure is a cornerstone of national development, underpinning mobility, economic activity, public safety, and territorial accessibility. The structural condition of pavement surfaces directly impacts vehicle performance, fuel efficiency, travel time reliability, and user safety. Poorly maintained roads lead to increased wear on vehicles and higher operating costs. In Europe, road maintenance alone can represent up to 40\% of total transport infrastructure spending, emphasizing the importance of targeted and timely maintenance efforts~\cite{schroten2019overview}.

For public administrations (PAs) managing extensive and aging road networks, early detection of surface anomalies is critical. Traditional visual inspection methods are labor-intensive, inconsistent, and lack scalability~\cite{koch2015review}. In Italy, for example, more than 250,000 kilometers of roads require regular assessment. The national road agency, ANAS, allocates over €1.5 billion annually to pavement rehabilitation~\cite{pompigna2022smart}, yet resource constraints continue to limit large-scale, proactive maintenance. As a result, AI-driven inspection systems are gaining traction as cost-effective, scalable alternatives~\cite{loprencipe2021validation}.

Recent advances in deep learning—particularly convolutional neural networks (CNNs)—have demonstrated strong performance in tasks such as crack classification, pothole detection, and texture anomaly recognition~\cite{zhang2020crackgan}. However, a key limitation of existing approaches is their primary focus on image-level classification, rather than on precise localization of defects. For real-world deployment, especially in public infrastructure management, simply knowing that a defect exists is insufficient. High-resolution, pixel-level localization is essential for prioritizing maintenance, estimating damage extent, and planning repairs efficiently.

\begin{figure}[tb]
  \centering
  \includegraphics[width=\textwidth]{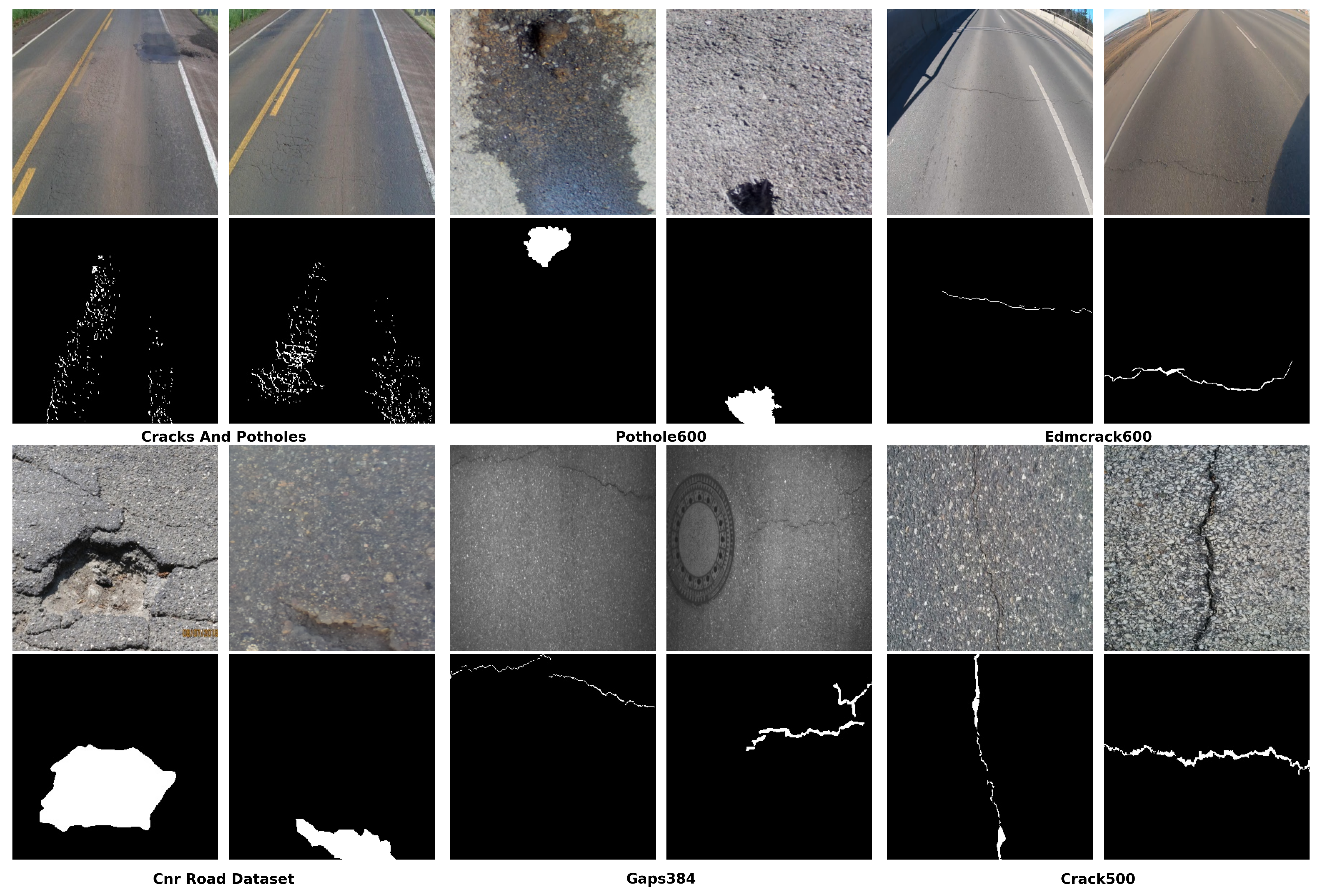}
  \caption{Diverse examples of real-world pavement defects from six benchmark datasets used in our experiments. Each pair shows an input image (top) and its corresponding ground truth mask (bottom). The datasets include a range of defect types and appearances: surface cracking and patching in Cracks and Potholes, localized potholes in Pothole600, fine linear cracks in Edmcrack600 and Gaps384, irregular surface damage in CNR Road, and dense crack patterns in Crack500. This visual diversity highlights the challenges of consistent defect detection across datasets.
  }
  \label{fig:defects}
\end{figure}

Additionally, several practical challenges persist. First, pre-trained models often struggle with domain shifts when applied to real pavement data. Second, the imbalance between abundant normal samples and limited defect samples reduces training effectiveness. Third, the visual diversity of pavement conditions—across materials, lighting, weather, and imaging perspectives—adds noise and complexity to feature learning. These issues contribute to false positives, missed detections, and unreliable predictions—outcomes that are costly and dangerous in practice~\cite{aqeel2025CoMet, garilli2024uav}. Figure~\ref{fig:defects} illustrates the visual diversity of real-world pavement defects across several benchmark datasets, including cracks, potholes, surface wear, and texture anomalies. These examples highlight the complexity of the task and the need for models that can generalize well across different defect types, scales, and appearances.

To address these challenges, we propose \methname, a novel framework that shifts the focus from simple defect classification to accurate spatial anomaly localization. The framework is designed to handle both the variability of real-world road conditions and the limitations of existing datasets. Our approach integrates synthetic anomaly generation using diffusion models with a dual-adaptor architecture that enhances feature learning for both normal and anomalous patterns.

Our main contributions are as follows:
\begin{itemize}
    \item A dual-adaptor architecture that bridges the domain gap between pre-trained features and pavement-specific representations, using separate pathways for normal and anomalous samples to improve discriminative power. 
    \item Integration of a latent diffusion model for generating diverse, realistic synthetic anomalies guided by text prompts and spatial masks, helping address the scarcity of annotated defect data. 
    \item A streamlined inference pipeline that maintains computational efficiency while delivering high-resolution anomaly localization across challenging, real-world datasets. 
\end{itemize}

\section{Related Work}
\label{sec:related}

Pavement defect detection has evolved from rule-based image processing to deep learning-driven approaches. Early methods relied on handcrafted features like Gabor filters and morphological operations~\cite{koch2015review}, but lacked robustness under real-world variations. Classical machine learning models (e.g., SVMs, Random Forests) improved performance by learning from labeled data, yet still depended on manual feature design.

Deep learning, particularly Convolutional Neural Networks (CNNs), marked a turning point by enabling end-to-end learning from raw imagery. Architectures such as U-Net~\cite{jenkins2018deep} and CrackGAN~\cite{zhang2020crackgan} brought pixel-level precision to defect localization, while real-time detectors like YOLO~\cite{maeda2018road} enabled practical deployment. More recently, hybrid models combining CNNs and transformers~\cite{jiang2025yolov5s} have improved context modeling, benefiting detection of subtle or large-spanning defects.

Surface defect detection, now dominated by CNNs, remains critical across industrial applications. Recent work has focused on making these models more robust and generalizable under real-world conditions~\cite{aqeel2025CoMet, aqeel2024meta}. Self-supervised learning approaches~\cite{aqeel2024self, Aqeel_2025} aim to improve defect detection without relying heavily on labeled datasets. By leveraging pretext tasks and unsupervised refinement strategies, these methods can identify subtle surface anomalies across varied domains.

Diffusion-based augmentation has shown promise in improving model performance under distribution shifts~\cite{capogrosso2024diffusion}. By generating realistic in-distribution samples, such approaches help mitigate overfitting and improve defect generalization, particularly relevant for surface inspection scenarios where data imbalance and domain variability are major obstacles.

Synthetic data generation has gained traction as a response to data scarcity in defect detection. GANs~\cite{kyslytsyna2021road} were initially adopted for augmentation, while diffusion models have emerged as a more stable and expressive alternative for generating high-quality defect samples~\cite{zhang2024crack}. These approaches enable the creation of diverse training examples covering a wider range of defect appearances and environmental conditions.

In parallel, domain adaptation and transfer learning strategies have addressed the mismatch between training distributions and deployment scenarios~\cite{he2022pavement}. These techniques help models maintain performance when faced with new pavement types, lighting conditions, or imaging systems not represented in the original training data. Collectively, these advancements represent a shift toward more data-efficient and adaptable solutions~\cite{girella2024leveraging}.

Despite these advances, challenges like domain shift, class imbalance, and scale variation persist. Our proposed framework, \methname, builds on these insights by combining diffusion-based synthetic anomaly generation with a dual-adaptor architecture, offering improved performance in both classification and localization tasks under diverse real-world conditions.

\section{\methname Pipeline}
\label{sec:method}

\begin{figure}[tb]
  \centering
  \includegraphics[width=1.0\textwidth]{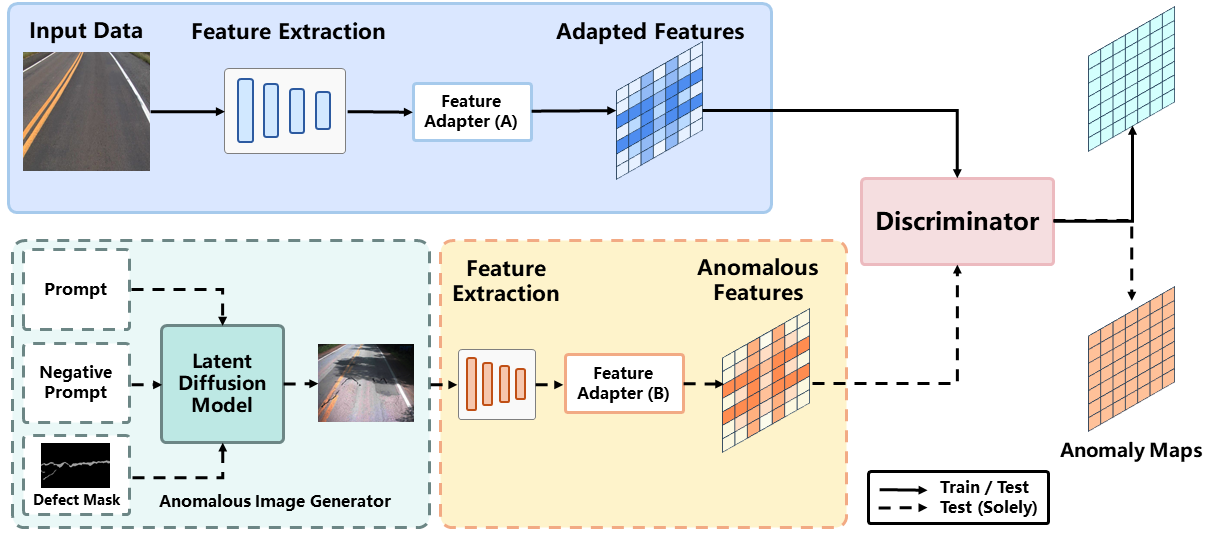}
  \caption{Overview of the proposed \methname architecture for pavement defect detection. The top pathway handles normal samples: defect-free road images are passed through a pre-trained feature extractor and then adapted via Feature Adapter (A) to generate domain-specific normal features. The bottom pathway generates synthetic anomalies using a latent diffusion model conditioned on prompts, negative prompts, and defect masks. These anomalous images are then processed through the same feature extraction pipeline and Feature Adapter (B) to obtain anomalous features. During training, the Discriminator learns to differentiate between normal and anomalous features. At test time, only the upper pathway is used to produce anomaly maps via the Discriminator.
  }
  \label{fig:method}
\end{figure}

The \methname framework is introduced in this section. As illustrated in Figure~\ref{fig:method}, \methname consists of a Feature Extractor, dual Feature Adaptors (A and B), a Latent Diffusion-based Anomalous Image Generator, and a Discriminator. The framework operates with a streamlined single-flow architecture during inference. These modules will be described below in sequence.

\subsection{Feature Extractor}

The Feature Extractor obtains local features through a multi-scale approach as in~\cite{roth2022towards}. We denote the training set and test set as $\mathcal{X}_{\text{train}}$ and $\mathcal{X}_{\text{test}}$. For any image $\mathbf{x}_i \in \mathbb{R}^{H \times W \times 3}$ in $\mathcal{X}_{\text{train}} \cup \mathcal{X}_{\text{test}}$, the pre-trained backbone network $\Phi$ extracts features from different hierarchical levels. We define $\mathcal{L}$ as the subset of selected hierarchical levels. The feature map from level $l \in \mathcal{L}$ is denoted as $\Phi_{l,i} \sim \Phi_l(\mathbf{x}_i) \in \mathbb{R}^{H_l \times W_l \times C_l}$, where $H_l$, $W_l$, and $C_l$ represent the height, width, and channel dimensions. For an entry $\Phi_{l,i}^{h,w} \in \mathbb{R}^{C_l}$ at location $(h, w)$, its neighborhood with patchsize $p$ is defined as:
\begin{equation}
\mathcal{N}_p^{(h,w)} = \left\{(h', w') \mid h' \in [h - \lfloor p/2 \rfloor, \ldots, h + \lfloor p/2 \rfloor],\ w' \in [w - \lfloor p/2 \rfloor, \ldots, w + \lfloor p/2 \rfloor]\right\}
\tag{1}
\end{equation}

Aggregating the features within the neighborhood $\mathcal{N}_p^{h,w}$ with aggregation function $f_{\text{agg}}$ (using adaptive average pooling) yields the local feature $\mathbf{z}_{l,i}^{h,w}$:
\begin{equation}
\mathbf{z}_{l,i}^{h,w} = f_{\text{agg}}\left(\left\{\Phi_{l,i}^{h',w'} \mid (h', w') \in \mathcal{N}_p^{h,w} \right\}\right)
\tag{2}
\end{equation}

To combine features $\mathbf{z}_{l,i}^{h,w}$ from different hierarchies, all feature maps are linearly resized to the same dimensions $(H_0, W_0)$. Concatenating the feature maps channel-wise produces the feature map $\mathbf{o}_i \in \mathbb{R}^{H_0 \times W_0 \times C}$:
\begin{equation}
\mathbf{o}_i = f_{\text{cat}}\left(\left\{\text{resize}(\mathbf{z}_{l',i}, (H_0, W_0)) \mid l' \in \mathcal{L} \right\}\right)
\tag{3}
\end{equation}

We define $\mathbf{o}_i^{h,w} \in \mathbb{R}^C$ as the entry of $\mathbf{o}_i$ at location $(h,w)$ and simplify the expression as:
\begin{equation}
\mathbf{o}_i = \mathcal{F}_{\Phi}(\mathbf{x}_i)
\tag{4}
\end{equation}
where $\mathcal{F}_{\Phi}$ represents the Feature Extractor.

\subsection{Feature Adaptors}

To adapt features to the target domain of pavement surfaces, we employ two distinct Feature Adaptors. Feature Adaptor A, denoted as $\mathcal{G}_A$, processes features from normal images:
\begin{equation}
\mathbf{q}_{h,w}^i = \mathcal{G}_A(\mathbf{o}_{h,w}^i)
\tag{5}
\end{equation}

For the anomalous images generated by the Latent Diffusion Model, we utilize Feature Adaptor B, denoted as $\mathcal{G}_B$. After extracting features from the synthetic anomalous images using the same Feature Extractor:
\begin{equation}
\mathbf{o}_a^i = \mathcal{F}_{\Phi}(\mathbf{i}_a^i)
\tag{6}
\end{equation}

These features are processed through Feature Adaptor B:
\begin{equation}
\mathbf{q}_{h,w}^{i-} = \mathcal{G}_B(\mathbf{o}_a^{i,h,w})
\tag{7}
\end{equation}

This separate adaptor pathway for anomalous features allows the framework to learn distinct representations for normal and defective pavement regions. Both Feature Adaptors A and B share the same architectural design, consisting of fully-connected layers, but maintain separate parameters to specialize in their respective domains. Experimental results demonstrate that this dual-adaptor approach more effectively differentiates between normal and anomalous features compared to using a single adaptor for both types of features.

\subsection{Latent Diffusion-based Anomalous Image Generator}

The Latent Diffusion Model (LDM)~\cite{ho2020denoising, girella2024leveraging} generates realistic pavement anomalies by leveraging a diffusion process that operates in a lower-dimensional latent space rather than directly in pixel space. In this paper, we generate images using DIAG~\cite{girella2024leveraging} for its ability to adapt to new textures and to cope with different surface defects. To generate an anomalous image $\mathbf{i}_a$, the process begins with a defect-free pavement image, a textual anomaly description, and a location mask, forming the triplet $(\mathbf{i}_n, \mathbf{d}_a, \mathbf{m}_a)$. The text-conditioned LDM performs inpainting on image $\mathbf{i}_n$ using the mask $\mathbf{m}_a$.

Given a set of defect-free pavement samples $\mathcal{I}_n$, the framework incorporates textual descriptions $\mathcal{D}_a$ of pavement anomalies (cracks, potholes, raveling, etc.). Regions where these anomalies may realistically appear are designated through a set of binary masks $\mathcal{M}_a$. The LDM, conditioned on this information, inpaints plausible anomalies onto the defect-free samples. The output $\mathbf{i}_a$ represents an anomalous version of $\mathbf{i}_n$, with a realistic defect inpainted in the masked region $\mathbf{m}_a$. This process can be repeated with different parameters to generate a diverse set of anomalous images $\mathcal{I}_a$ for training.

\subsection{Discriminator}

The Discriminator $\mathcal{D}_{\psi}$ functions as a normality estimator, calculating a normality score at each spatial location $(h, w)$. It processes both normal features $\{\mathbf{q}^i \mid \mathbf{x}_i \in \mathcal{X}_{\text{train}}\}$ from Feature Adaptor A and anomalous features $\{\mathbf{q}^{i-}\}$ from Feature Adaptor B during training. The Discriminator architecture employs a 2-layer MLP that outputs a scalar normality value $\mathcal{D}_{\psi}(\mathbf{q}_{h,w}) \in \mathbb{R}$.

\subsection{Loss Function and Training}

The training employs a truncated $\ell_1$ loss formulation:
\begin{equation}
\ell_{h,w}^i = \max\left(0, \tau_+ - \mathcal{D}_{\psi}(\mathbf{q}_{h,w}^i)\right) + \max\left(0, -\tau_- + \mathcal{D}_{\psi}(\mathbf{q}_{h,w}^{i-})\right)
\tag{8}
\end{equation}
where $\tau_+$ and $\tau_-$ represent threshold values set to 0.5 and $-0.5$ respectively. The overall training objective is:
\begin{equation}
\mathcal{L} = \min_{A, B, \psi} \sum_{\mathbf{x}_i \in \mathcal{X}_{\text{train}}} \sum_{h,w} \frac{\ell_{h,w}^i}{H_0 \cdot W_0}
\tag{9}
\end{equation}
where $A$ and $B$ are the parameters of Feature Adaptors A and B respectively. The performance of this loss function is evaluated against standard cross-entropy loss in the experiments section. 

\subsection{Inference and Scoring Function}

During inference, the Latent Diffusion-based Anomalous Image Generator and Feature Adaptor B are not used. For each test image $\mathbf{x}_i \in \mathcal{X}_{\text{test}}$, features are extracted through the Feature Extractor $\mathcal{F}_{\Phi}$ and adapted via Feature Adaptor A $\mathcal{G}_A$ to obtain features $\mathbf{q}_{h,w}^i$ as in Equation~(5). The anomaly score is calculated by the Discriminator $\mathcal{D}_{\psi}$:
\begin{equation}
\mathbf{s}_{h,w}^i = -\mathcal{D}_{\psi}(\mathbf{q}_{h,w}^i)
\tag{10}
\end{equation}

The anomaly map for localization is defined as:
\begin{equation}
\mathbf{S}_{AL}(\mathbf{x}_i) := \left\{\mathbf{s}_{h,w}^i \mid (h, w) \in H_0 \times W_0\right\}
\tag{11}
\end{equation}

This map is interpolated to match the input resolution and smoothed with a Gaussian filter ($\sigma = 4$). The final anomaly detection score for each image is determined by taking the maximum value from the anomaly map:
\begin{equation}
\mathbf{S}_{AD}(\mathbf{x}_i) := \max_{(h,w) \in H_0 \times W_0} \mathbf{s}_{h,w}^i
\tag{12}
\end{equation}

\section{Experimental Results}
\label{sec:experiment}

\subsection{Datasets}

We evaluate our method on six public road damage datasets, each offering a unique combination of image characteristics and defect types:

\begin{itemize} \item \textbf{Crack500}~\cite{zhang2016road}: 500 high-resolution images ($2000 \times 1500$) captured via smartphone, annotated for cracks.

\item \textbf{GAPs384}~\cite{eisenbach2017get}: 384 grayscale images ($1920 \times 1080$) manually selected to focus on crack detection.

\item \textbf{EdmCrack600}~\cite{mei2020densely}: 600 pixel-level annotated images of road cracks from urban roads in Edmonton, Canada.

\item \textbf{Pothole-600}~\cite{fan2020we}: 600 RGB images ($400 \times 400$) annotated for potholes, with accompanying disparity maps and masks.

\item \textbf{CPRID}~\cite{passos2020cprid}: 2,235 images ($1024 \times 640$) from Brazilian highways, labeled for both cracks and potholes.

\item \textbf{CNR Road}~\cite{thompson2022shrec}: 20 high-resolution web images with detailed annotations of potholes. \end{itemize}

These datasets span various regions, imaging methods, and damage types, forming a comprehensive benchmark for evaluating road surface anomaly detection.

\subsection{Implementation Details}

We implemented our model using the PyTorch framework and trained it on an NVIDIA RTX 4090 GPU for efficient training and inference. We use a pre-trained WideResNet-50 as the Feature Extractor, with features extracted from multiple intermediate layers and aggregated into a 1536-dimensional representation. Feature Adaptors A and B are fully connected layers without bias, sharing architecture but using separate parameters. Anomalous images are generated using a latent diffusion model guided by text prompts and spatial masks, then passed through the same feature pipeline to obtain anomalous features. The Discriminator is a two-layer MLP with batch normalization and leaky ReLU (slope 0.2). We train the model using Adam with learning rates of 0.0001 for the adaptors and 0.0002 for the discriminator, a weight decay of 0.00001, batch size 16, and 60 training epochs.

\subsection{Evaluation Metrics}
We evaluate performance using a comprehensive set of metrics to assess both classification and localization capabilities. For classification, we report Precision, Recall, Macro-F1, and AUROC—including both image-level (I-AUROC) and pixel-level (P-AUROC) variants. For localization and segmentation quality, we include mean Average Precision (mAP), Intersection over Union (IoU), and Average Precision (AP) where applicable. These metrics provide a balanced view of the model's accuracy, generalization, and ability to precisely identify defect regions.

\vspace{-1em}
\begin{table}[t!]
\centering
\caption{Performance comparison of different methods across multiple datasets. Best results for each metric are highlighted in \textbf{bold}.}
\scalebox{0.85}{
\begin{tabular}{llcccccccc}
\toprule
\textbf{Dataset} & \textbf{Method} & \textbf{P.} & \textbf{R.} & \textbf{M.-F1} & \textbf{mAP} & \textbf{IoU} & \textbf{AP} & \textbf{I/A} & \textbf{P/A} \\
\midrule
\multirow{2}{*}{CNR Road}
  & Eisenbach~\cite{eisenbach2017get} & 0.85 & \textbf{0.91} & 0.85 & 0.79 & 0.70 & \textbf{0.85} & 0.77 & 0.81 \\
  & \methname                         & \textbf{0.89} & 0.87 & \textbf{0.88} & \textbf{0.83} & \textbf{0.76} & 0.81 & \textbf{0.84} & \textbf{0.87} \\
\midrule
\multirow{3}{*}{Crack500}
  & Liu~\cite{liu2019deepcrack}       & 0.85 & 0.85 & 0.86 & 0.83 & 0.74 & --- & 0.73 & 0.71 \\
  & Yang~\cite{yang2019feature}       & \textbf{0.93} & 0.85 & 0.88 & \textbf{0.90} & 0.73 & --- & 0.74 & 0.75 \\
  & \methname                         & 0.91 & \textbf{0.90} & \textbf{0.91} & 0.88 & \textbf{0.79} & \textbf{0.89} & \textbf{0.79} & \textbf{0.81} \\
\midrule
\multirow{2}{*}{Cracks \& Potholes}
  & Maeda~\cite{maeda2018road}        & \textbf{0.89} & 0.80 & 0.84 & \textbf{0.84} & 0.68 & \textbf{0.85} & 0.79 & 0.78 \\
  & \methname                         & 0.87 & \textbf{0.89} & \textbf{0.88} & 0.82 & \textbf{0.74} & 0.83 & \textbf{0.82} & \textbf{0.80} \\
\midrule
\multirow{2}{*}{EDM Crack}
  & Zhang~\cite{zhang2017automated}   & \textbf{0.86} & 0.78 & 0.82 & \textbf{0.83} & 0.67 & \textbf{0.84} & 0.75 & 0.74 \\
  & \methname                         & 0.82 & \textbf{0.83} & \textbf{0.84} & 0.79 & \textbf{0.72} & 0.80 & \textbf{0.76} & \textbf{0.81} \\
\midrule
\multirow{2}{*}{GAPS384}
  & Eisenbach~\cite{eisenbach2017get} & 0.86 & \textbf{0.91} & \textbf{0.88} & \textbf{0.88} & 0.71 & \textbf{0.89} & 0.80 & \textbf{0.82} \\
  & \methname                         & \textbf{0.90} & 0.88 & 0.87 & 0.85 & \textbf{0.78} & 0.86 & \textbf{0.84} & 0.78 \\
\midrule
\multirow{2}{*}{Pothole600}
  & Dhiman \& Klette~\cite{dhiman2019pothole} & \textbf{0.89} & 0.78 & 0.83 & \textbf{0.83} & 0.67 & \textbf{0.84} & 0.72 & 0.75 \\
  & \methname                                & 0.88 & \textbf{0.86} & \textbf{0.87} & 0.81 & \textbf{0.75} & 0.82 & \textbf{0.79} & \textbf{0.83} \\
\bottomrule
\end{tabular}
}
\label{tab:comprehensive}
\end{table}

\subsection{Quantitative Results}
\methname demonstrates superior performance across all six benchmark datasets as shown in Table~\ref{tab:comprehensive}, confirming its robustness to diverse pavement types and defect characteristics. On Crack500, it achieves the highest Macro-F1 (0.91) and Recall (0.90), with a strong P-AUROC of 0.73—critical for early detection where missing subtle cracks can accelerate infrastructure deterioration. For CNR Road, \methname reports superior Precision (0.89) and Macro-F1 (0.88), along with the highest I-AUROC (0.82), indicating reliable discrimination between normal and defective pavements. When handling the multi-defect Cracks \& Potholes dataset, \methname outperforms baselines in both Recall (0.89) and Macro-F1 (0.88), while delivering the highest P-AUROC (0.80), demonstrating adaptability to scenes with mixed damage categories. On the challenging EDM Crack dataset, characterized by fine-scale crack structures, \methname improves IoU from 0.67 to 0.72 compared to baselines and achieves a P-AUROC of 0.81.

For GAPS384, \methname attains the highest IoU (0.78) and mAP (0.85), confirming accurate defect localization even in visually complex road surfaces. On Pothole600, containing large, irregular defects, it maintains high P-AUROC (0.83) with an IoU of 0.75, alongside balanced Precision and Recall metrics resulting in a strong Macro-F1 of 0.87. The consistent excellence in both detection metrics (Macro-F1, AUROC) and localization metrics (IoU, mAP) across all datasets demonstrates \methname's superior generalization capabilities, from fine cracks to extensive potholes, making it a robust solution for real-world pavement monitoring applications.

\subsection{Synthesized Anomalies}

\begin{figure}[tb]
  \centering
  \includegraphics[width=\textwidth]{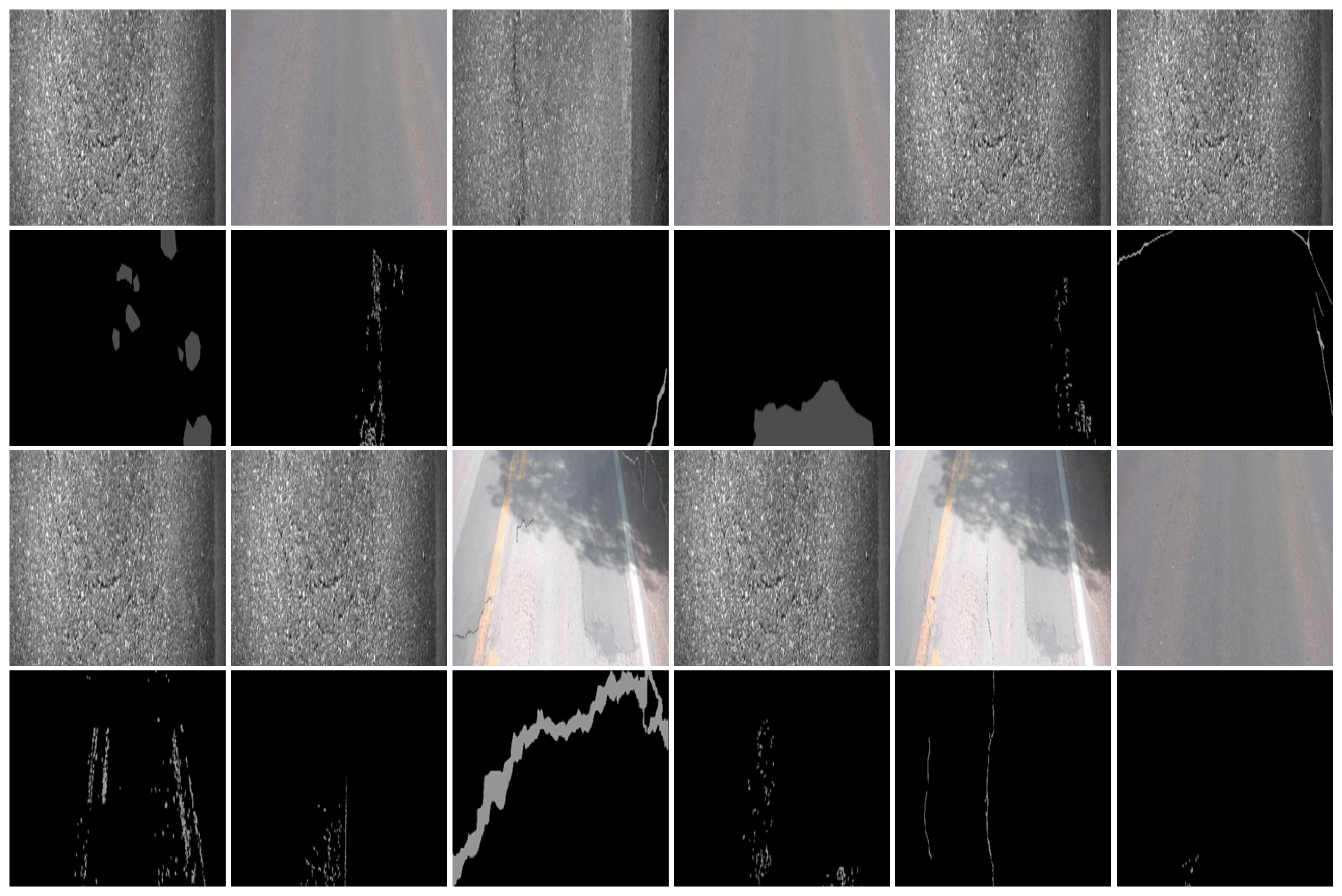}
  \caption{Examples of synthesized pavement anomalies generated by the latent diffusion model. The samples show a range of defect types—including cracks, surface erosion, and patch damage—inpainted onto clean road images using textual prompts and spatial masks. These synthetic anomalies are used during training to extract anomalous features for the discriminator.
  }
  \label{fig:augment}
\end{figure}

Figure~\ref{fig:augment} shows a selection of anomalous pavement images generated by our diffusion-based anomaly synthesis pipeline. These samples were created by inpainting realistic defects, such as cracks, patches, and surface damage, onto clean road images using textual prompts and binary location masks. The visual diversity in shape, texture, and scale mirrors real-world defect patterns, which helps the model generalize effectively during training. These images serve as the input for extracting anomalous features via the backbone network and are subsequently processed through Feature Adaptor B in our training pipeline.

\section{Conclusion}
We introduced \methname, a diffusion-based framework for pavement defect detection that combines anomaly generation with dual-adaptor feature learning. By leveraging latent diffusion to synthesize diverse pavement anomalies, our approach addresses the limited availability of labeled defect data. The dual feature adaptors enable domain-specific feature alignment, improving defect localization and classification. Experiments across six benchmark datasets demonstrate that \methname consistently outperforms existing methods in both detection and localization tasks. 
Results confirm strong generalization across various road surfaces and defect types, while qualitative samples validate the realism of synthesized anomalies. \methname provides an effective solution for pavement monitoring when annotated data are limited or diverse defect types are expected.

\section*{Acknowledgements}
This study was carried out within the PNRR research activities of the consortium iNEST (Interconnected North-Est Innovation Ecosystem) funded by the European Union Next-GenerationEU (Piano Nazionale di Ripresa e Resilienza (PNRR) – Missione 4 Componente 2, Investimento 1.5 – D.D. 1058  23/06/2022, ECS\_00000043).

\par\vfill\par

%
%
\bibliographystyle{splncs04}
\bibliography{main}
\end{document}